\documentclass[10pt,twocolumn,letterpaper]{article}

\usepackage{cvpr}
\usepackage{times}
\usepackage{epsfig}
\usepackage{graphicx}
\usepackage{amsmath}
\usepackage{amssymb}
\usepackage{multirow}
\usepackage{multicol}  
\usepackage{booktabs}
\usepackage{floatrow}
\usepackage{extarrows}
\usepackage{colortbl} 
\usepackage{arydshln} 

\usepackage[T1]{fontenc}
\usepackage[utf8]{inputenc}
\usepackage{authblk}

\usepackage[breaklinks=true,bookmarks=false]{hyperref}

\cvprfinalcopy 

\def\cvprPaperID{****} 

\begin{document}

\title{Class-Conditional Domain Adaptation on Semantic Segmentation}

\author{{Yue Wang}$^1$,
{Yuke Li}$^2$,
{James H. Elder}$^2$,
{Runmin Wu}$^1$,
{Huchuan Lu}$^1$\\

$^1$Dalian University of Technology, China,\\
$^2$York University, Canada\\
}

\maketitle

\begin{abstract}
Semantic segmentation is an important sub-task for many applications, but 
pixel-level ground truth labeling is costly and there is a tendency to overfit
the training data, limiting generalization.  Unsupervised domain adaptation
can potentially address these problems, allowing systems trained on labelled datasets from one
or more source domains (including less expensive synthetic domains) to be adapted
to novel target domains.  
The conventional approach is to automatically align the representational distributions
of source and target domains. 
One limitation of this approach is that it tends to disadvantage lower probability classes.
We address this problem by introducing a Class-Conditional Domain Adaptation method (CCDA).
It includes a class-conditional multi-scale discriminator and the class-conditional loss.
This novel CCDA method encourages the network to shift the domain in a class-conditional manner,
and it equalizes loss over classes.
We evaluate our CCDA method on two transfer tasks and demonstrate performance comparable to state-of-the-art methods.
\end{abstract}

\section{Introduction}
Semantic segmentation is an important visual scene understanding task with wide application, especially in 
autonomous and assisted vehicle systems. 
Recent deep network approaches (e.g., ~\cite{long2015fully, zhao2017pyramid, chen2018deeplab}) have
achieved impressive results, but 
require large training datasets with precise pixel-level ground-truth annotation 
and do not generalize well
over large domain shifts in viewpoint, lighting, etc.~\cite{yao2015semi}

These issues can potentially be addressed by unsupervised domain adaptation methods 
that attempt to identify and correct for a shift in the appearance of the visual input from source to target domains.  
A successful domain adaptation method will not only improve generalization, 
but allow larger and more easily obtained synthetic ground truth to be used for training, 
even if it does not perfect represent the appearance of real scenes.
The well-generalized model can be trained with synthetic image dataset which has access to ground-truth label,
and real-world image dataset whose ground-truth label remains unknown,
to avoid labor costing and high time-consuming annotation job.
The synthetic image dataset with ground-truth label is named as source domain, 
while the real-world image dataset without ground-truth label is named as target domain.

A common approach to solve the "domain shift" problem for deep network systems 
is to modify the weights of the network to render representations produced by the network for target domain vectors more similar to representations produced for source domain vectors.
By further minimizing the distance between distributions of certain representations from both domains,
a well-generalized model can be obtained.
Some papers have focused on representations in the prediction space~\cite{tsai2018learning, vu2019advent}
while others have focused on representations in feature (latent) space~\cite{hoffman2016fcns, chen2017no, luo2019significance}.
Representational dissimilarity can be assessed using correlation distances~\cite{sun2016return} or maximum mean discrepancy~\cite{geng2011daml}.
However, more recent work has tended to focus on generative adversarial methods~\cite{goodfellow2014generative} 
for unsupervised domain adaptation.
This adversarial principle becomes prominent since it 
achieves promising result for pixel-level prediction tasks~\cite{hoffman2016fcns, hoffman2018cycada, tsai2018learning, vu2019advent, saito2018maximum, luo2019taking}.

One limitation of prior work on unsupervised domain adaptation for semantic segmentation is that 
domain adaptation tends to be more effective for more frequent classes~\cite{hoffman2016fcns, tsai2018learning}.
An underlying tendency can be observed that
representations on classes with higher frequency can be easily extracted and adapted,
while certain classes with lower frequency are inclined to be failed.
For driving datasets such as 
Cityscapes~\cite{cordts2016cityscapes}, 
adaptation works fairly well for dominant classes such as road, car, buildings, vegetation, and sky, 
but less well for infrequent classes such as sign or bicycle.

To address this issue, we propose a novel Class-Conditional Domain Adaptation method (CCDA).
It consists of a class-conditional multi-scale discriminator,
and a class-conditional loss function for both segmentation and adaptation.
The basic idea of our class-conditional multi-scale discriminator is to measure the alignment of 
feature-level representations at both fine and coarse spatial scales. 
The fine-scale branch is to evaluate the adaptation on pixel-level,
while in particular, for each coarse-scale patch in the image, 
the loss is weighted equally over all classes occurring (or estimated to occur) within the patch, 
regardless of the number of pixels associated with each class.  
Our class-conditional multi-scale discriminator not only encourages the network to realign representations of 
pixels belonging to the same class in a consistent, class-conditional way,
but also provides equal attention on each class occurred in one patch.
Meanwhile, the design of the class-conditional loss function is also to assist the network 
to evaluate the performance of both segmentation and adaptation on each class fairly.

In summary, our proposed CCDA method makes three main contributions:
\begin{itemize}
\item {We proposed a novel class-conditional multi-scale discriminator that allows class-conditional domain shift to be learned.  }
\item {By equalizing the class-conditional loss for both segmentation and adaptation, 
we further improve the performance for less frequent classes.}
\item {We demonstrate that our method achieves comparable performance to state-of-the-art algorithms 
on two semantic segmentation domain adaptation scenarios.}
\end{itemize}

\section{Related Work}

While there  has been substantial progress on domain adaptation for image classification~\cite{tzeng2017adversarial, ganin2015unsupervised, long2015learning, long2016unsupervised, hu2018duplex, zhang2019domain, xie2018learning, pan2019transferrable, ganin2016domain}, 
pixel-level tasks are more challenging due to the more direct dependence on local appearance.  Nevertheless,  increasing activity in autonomous vehicle applications has driven interest in domain adaptation for pixel-level segmentation of road scenes~\cite{hoffman2016fcns, zhang2017curriculum, chen2017no, vu2019advent, zou2018unsupervised, hong2018conditional, tsai2018learning, murez2018image, sankaranarayanan2018learning, zhang2018fully, zhu2018penalizing, chen2018road, luo2019taking}.

The most popular current approach relies on adversarial learning, where a discriminator is employed to align source and target representations either at the prediction-level~\cite{tsai2018learning, vu2019advent}
or the feature-level~\cite{hoffman2016fcns, chen2017no, luo2019significance}.
In~\cite{tsai2018learning}, Tsai \emph{et al.} first provide a prediction-level representation alignment with GAN network
for domain adaptation on semantic segmentation.
Vu \emph{et al.}~\cite{vu2019advent} then employe an entropy minimization technique during adversarial learning to improve domain adaptation at the prediction level and 
Luo \emph{et al.}~\cite{luo2019significance} use an information bottleneck approach to more fully remove task-independent information from feature representations.
Co-training adaptation using multi-view learning 
has also been employed~\cite{saito2018maximum, saito2017adversarial, luo2019taking}.

\begin{figure*}[ht]
\centering
\includegraphics[width=1\textwidth]{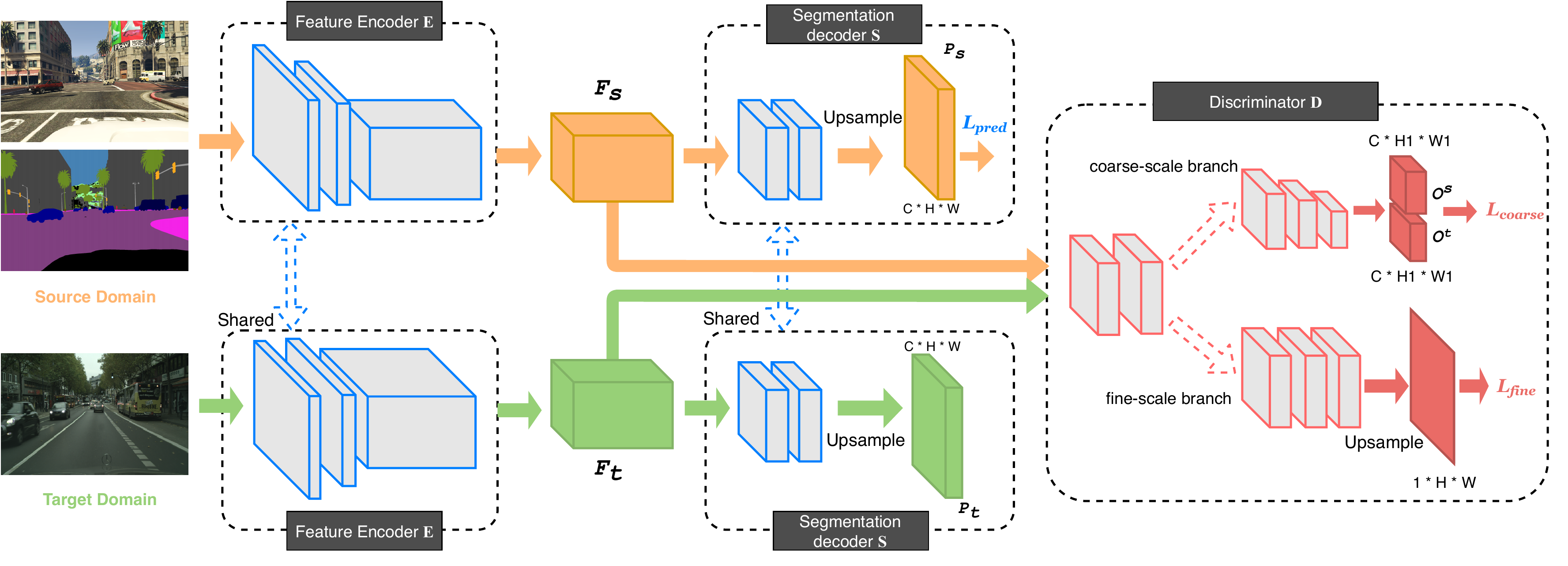}
 \caption{Overview of our proposed Class-Conditional Domain Adaptation.}
\label{fig:structure}
\end{figure*}

Approaches like pixel-level adaptation and self-training provide different directions for the process of domain adaptation,
and can be combined with the above representation adaptation methods.
The pixel-level adaptation approach is to view domain adaptation in part as a style transfer problem.  
In this approach, images from one domain are transformed to have the `style' or appearance of images from another domain, while preserving the original `content' of the image from the first domain~\cite{zhang2018fully, sankaranarayanan2018learning, li2019bidirectional}.
The self-training approach is to alternatively select unlabelled target samples with higher prediction probability
and utilize them with their predictions as pseudo ground-truth labels during training while updating the learnt model
\cite{zou2018unsupervised, chen2019domain}.


Techniques like~\cite{chen2017no,  luo2019taking, du2019ssf, tsai2019domain} 
tend to boost domain adaptation performance for some classes or regions of the image more than others, 
suggesting that a class- or region-conditioned domain adaptation approach may be required to achieve good adaptation over all classes.  
In~\cite{chen2017no, du2019ssf}, an adversarial system is employed to train distinct domain classifiers for each segmentation class.  
Luo \emph{et al.}~\cite{luo2019taking} instead use the disagreement between two classifiers to indicate the probability of incorrect representational alignment for each region of the image, increasing the weight of the adversarial loss for regions that appear to be poorly aligned.
Tsai \emph{et al.}~\cite{tsai2019domain} utilize the multiple modes of patch-level prediction with a more accurate classification for the category distribution, and apply the adaptation based on the representation of this patch classification. 
Meanwhile, for the self-training method., Zou \emph{et al.}~\cite{zou2018unsupervised} employ class-normalized confidence scores for pseudo ground-truth label selection to prevent the imbalanced selection of target domain samples on each class, which improves the performance on less frequent classes.  

The common drawback of the pervious adversarial learning adaptation methods is they neglect the imbalance frequency of different classes even though they consider the class information.
They fail to apply the equal attention on each class by not taking the class-based performance into account. 
Here we propose a class-conditional multi-scale discriminator
and a class-conditional loss function for both segmentation and adaptation.
By using the designed class label for each patch,
we allow the discriminator to consider class-conditional information for adaptation
equal to all classes.
The way of equalizing the loss over classes also improves the performance for lower frequent classes. 
Our method is more efficient than~\cite{chen2017no, du2019ssf} since we avoid training multi domain classifiers for each
class, and the multi-scale discriminator encourages to capture the domain shift and evaluate the adaptation in pixel-level
as well as patch-level.    

\section{Methods}

In this section, we present our proposed CCDA approach using class-conditional multi-scale discriminator
and our class-conditional loss function for segmentation and adaptation.
We begin by describing a basic structure for domain adaptation.
Then, we will explain in detail innovations for our class-conditional multi-scale discriminator,
and describe the design of class-based loss function for adaptation and segmentation components.
The overview of our proposed structure is showed in Figure \ref {fig:structure}.

\subsection{Basic Domain Adaptation Architecture}  \label{section: Basic Domain Adaptation Structure}

We apply an adversarial learning approach for our unsupervised domain adaptation on segmentation,
since it is the most explored way in this area. 
A basic structure consists of three modules:
a feature encoder $\mathbf E$, a segmentation decoder $\mathbf S$,
and a discriminator $\mathbf D$.
The image data consist of source domain data and target domain data.
Each source image $ I_s \in\mathcal{I}_s$ is paired with ground truth pixel-level segmentation labels
$Y_s\in \mathcal{Y}_s$.  Target images $I_t \in\mathcal{I}_t$ are assumed to have no ground truth data
available for training.  

Our goal is to train the feature encoder $\mathbf E$ and segmentation decoder $\mathbf S$ to output good prediction
$ P_t $  on target domain image.
This is achieved through two processes, one that trains $\mathbf E$ and $\mathbf S$ to 
output good segmentation prediction $ P_s $ for source image $I_s$ with associated label $ Y_s$,
and a second that uses the discriminator $\mathbf D$  to align the feature-level representations $ F_s $ and $ F_t $ 
output by the feature encoder $\mathbf E$ for the two domains.

The first (segmentation) process is trained by minimizing the segmentation cross-entropy loss: 
\begin{equation}
\label{segloss1}
\mathcal{L}_{seg}(\mathbf E, \mathbf S) = -\sum_{h=1,w=1}^{H,W} \sum_{c=1}^C Y_s^{(h, w, c)}log(P_s^{(h, w, c)})
\end{equation}
where $ H, W $ are the size of image, $ C $ is the number of semantic class.
$Y_s^{(h, w, c)}$ and $P_s^{(h, w, c)}$ are the ground truth and predicted states for Class $c$ at pixel $(h,w)$. 
$P_s = \mathbf S(F_s) = \mathbf S(\mathbf E(I_s))$ is the output of segmentation decoder $\mathbf S$.

The second (alignment) process is trained adversarially to generate domain-invariant features. 
Our discriminator module $\mathbf D$ tries to distinguish feature representations from source and target domains,
minimizing
\begin{equation}
\label{Dloss1}
\mathcal{L}_{D1}(\mathbf{D}) = \lambda_s \mathcal{L}_{bce}(\mathbf D(F_s), 0) + 
\lambda_t \mathcal{L}_{bce}(\mathbf D(F_t), 1)
\end{equation} 
where $\mathcal{L}_{bce}$ is the binary cross-entropy domain classification loss
since the output channel of this basic discriminator D is 1.
And source and target domain samples are assigned labels of 0 and 1, respectively.
Concurrently, the feature encoder $\mathbf E$ tries to confuse $\mathbf D$, minimizing
\begin{equation}
\label{advloss1}
\mathcal{L}_{adv1}(\mathbf E) = \lambda_s \mathcal{L}_{bce}(\mathbf D(F_s), 1) + 
\lambda_t \mathcal{L}_{bce}(\mathbf D(F_t), 0)
\end{equation} 

This adversarial learning process produces a rough alignment of features among all classes,
but tends to work less well for lower frequency classes that do not contribute 
substantially to the cross-entropy loss.  Also, since the feature map computed 
by our encoder is spatiotopic but reduced in resolution relative
to the input, the alignment achieved by our adversarial process is at
the specific intermediate scale of our feature map, which may not capture domain
shift at smaller or larger scales.  
These observations motivate our class-based multi-scale discriminator
and class-based loss function for segmentation and adaptation.

\subsection{Class-Conditional Multi-scale Discriminator}  \label{section: Weakly Class-based Multi-level Discriminator}

Our proposed class-conditional multi-scale discriminator is composed of fine-scale and coarse-scale branches (Fig. \ref{fig:structure}),
The fine-scale branch measures alignment at the pixel-level using the basic architecture with loss functions in Equations \ref{Dloss1} and \ref{advloss1}
and thus can capture spatially detailed domain shift phenomena.
The coarse-scale branch measures class-conditional alignment
at a scale that is coarser than the feature scale with the equal class information.  
We first describe how to perform this class conditioning
by explaining the coarse-scale class label, and then elaborate on our structure of the class-based coarse-scale discriminator
branch as well as the class-based fine-scale branch.

\subsubsection{Coarse-Scale Class Labels}
We define a coarse-scale class label $ W \in \{0,1\}^C$ that indicates the presence or absence of each class within a rectangular patch of the image. 
Note that a patch may contain multiple classes, this class label is not a one-hot label. 
For source images, $W$ is computed by analyzing the pixel-level ground-truth labels $Y_s$ within the image back-projection of a patch.
If any pixel within the back-projected region of the image has class $c$, we set $W^c=1$, otherwise we set $W^c=0$.

For target domain images, we do not have ground-truth labels.
Instead, we assign coarse-scale class labels based on the projected pixel-level predictions $P_t^c$ of our segmentation module $\mathbf S$ for the patch.
In particular, given a confidence threshold $th_w$, if $P_t^c> th_w$ for any pixel within a patch,
we set $W^c=1$, otherwise we set $W^c=0$.

Note that binarizing the patch-based class label $W$ has the effect of equalizing class frequencies
at the patch level:  $W^c=1$ whether the number of pixels with class $c$ the patch contains.  
This will have the benefit or boosting adaptation performance for less frequent classes
by apply an equal attention on all the classes a patch contains.

\subsubsection{Class-conditional Coarse-scale Branch}  \label{section: Class-conditional Coarse-scale Branch}
For the basic domain adaptation in Section \ref{section: Basic Domain Adaptation Structure}, 
the discriminator output is a scalar value indicating the domain of the input vector (in our case, 0 for source domain, 
1 for target domain).
In contrast, the output of our class-conditional coarse-scale discriminator branch 
consists of two vectors $O^s$ and $O^t$, each of length $C$.  
$O^s$ carries estimates of patch-level class labels for the source domain,  
an large value on $O^s_c$ indicates high confidence that the patch contains at least one pixel drawn from
the source domain and belonging to class $c$.  
And similarly, $O^t$ carries estimates of patch-level class labels for the target domain.

\begin{figure}[t] 
\centering
\includegraphics[width=1\textwidth]{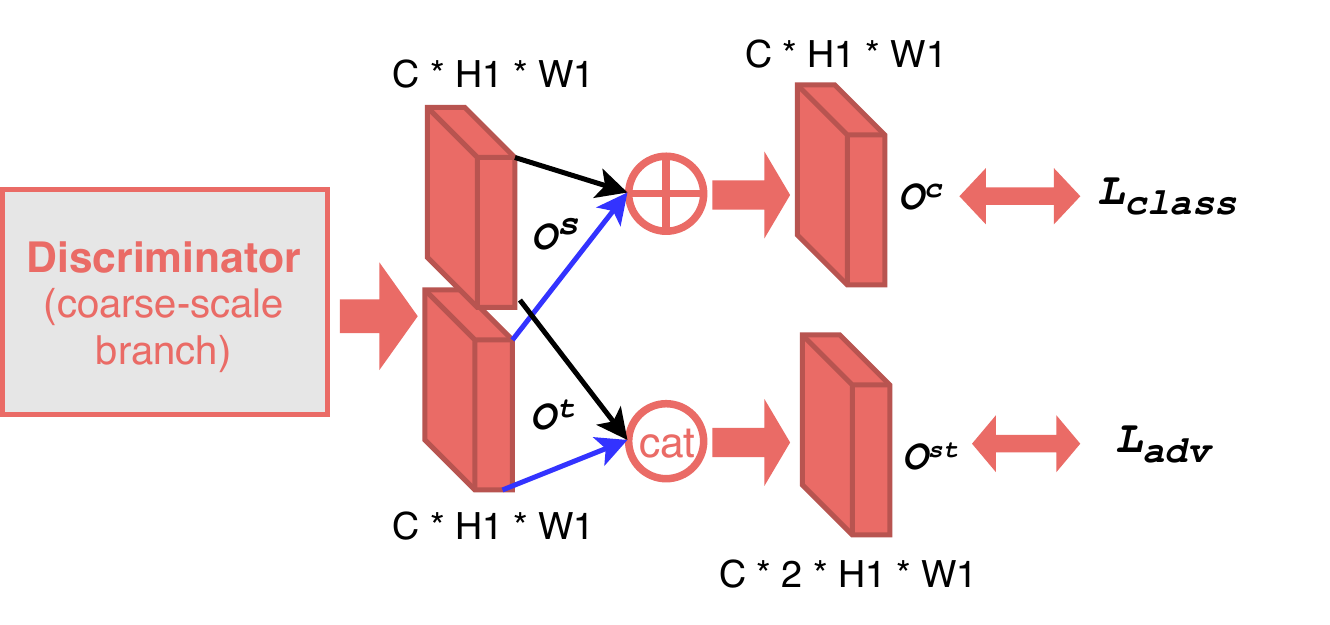}
 \caption{Details about the loss calculation in coarse-scale class-based branch in discriminator.}
\label{fig:adv_class}
\end{figure}

The advantage of this dual vector representation is that it allows us to multiplex 
both domain and class information, informing both an adversarial adaptation loss based on class
and a non-adversarial classification loss (Figure \ref{fig:adv_class}).  
In particular, to inform the non-adversarial classification loss, we form the
vector $O^c=\sigma\left(O^s+O^t\right)$,
where $\sigma(\cdot)$ is a sigmoid function apply for each class. 
And we calculate a classification loss with the binary cross-entropy loss as $\mathcal{L}_{bce}(O^c, W^c)$. 
Note that including this classification loss in the discriminator will encourage
a feature-level domain alignment that preserves segmentation class information.
Meanwhile, the designed class label $W$ insures the class information it preserves is equal to all classes,
and improve the performance on less frequent classes. 

To inform the adversarial adaptation loss, we form the $C\times2$ matrix $O^{st}=f\left(\left[O^s, O^t\right]\right)$,
where $f(\cdot)$ is the softmax operation over rows, normalizing the sum of $O^s_c$ and $O^t_c$ to 1 for every class
$c$.  
$O^{st}$ indicates the probability of the pixels belong to certain classes on a patch are from source domain or target
domain.
This normalization will tend to spread the loss more uniformly across classes. 
However, it should be noted that one patch may only contain certain classes instead of all $C$ classes,
so not all values on $C$ channels are valuable for calculating the loss function for domain label classification.
To form the final discriminator
domain adaptation loss, we sum the class-conditional loss over classes present in the patch by the classes one contains, 
weighting the sum
by the ground truth patch-level class labels $W_s$ for the source domain, 
and the predicted patch-level class predictions $O_t$ for the target domain.  
Combining with the non-adversarial classification loss, the total patch-level discriminator loss is:
{\setlength\abovedisplayskip{6pt}
\begin{equation}
\begin{aligned}
\label{Dloss_class}
\mathcal{L}_{D \_coarse}(\mathbf D) = \ &\lambda_c \mathcal{L}_{bce}(O^c, W^c)
\\ + &\lambda_s \sum_{c=1}^ C W_s^c[c] \mathcal{L}_{ce}(O_s^{st} [c], 0) 
\\ +  & \lambda_t \sum_{c=1}^ C O_t^c[c] \mathcal{L}_{ce}(O_t^{st} [c], 1) 
\end{aligned}
\end{equation} 
\setlength\belowdisplayskip{-6pt}}
where $O_s^{st}$ is the output $O_{st}$ for source domain images and $O_t^{st}$ is the output $O_{st}$ for source domain images.

The generative component of the adversarial loss is defined symmetrically:
\begin{small}
\begin{equation}
\begin{aligned}
\label{advloss_class}
\mathcal{L}_{adv \_coarse}(\mathbf E, \mathbf S) = \ &\lambda_c \mathcal{L}_{bce}(O^c, W^c)
\\ + &\lambda_s \sum_{c=1}^ C W_s^c[c] \mathcal{L}_{ce}(O_s^{st} [c], 1) 
\\ +  & \lambda_t \sum_{c=1}^ C O_t^c[c] \mathcal{L}_{ce}(O_t^{st} [c], 0) 
\end{aligned}
\end{equation} 
\end{small}
\subsubsection{Fine-Scale Class-Conditional Discriminator} \label{section: Class-based Loss for Adaptation}
The coarse-scale class-conditional adaptation can capture larger-scale domain shift effects
but may not capture shifts in finer detail.  For that purpose, we employ a fine-scale class-conditional
discriminator which can evaluate the adaptation on a pixel-level way.
It remains the scale of feature representations in this fine-scale discriminator branch and upsamples the output 
to produce a fine-scale domain classification $U_s$ and $U_t$ that match the size of the original input.
For this fine-scale discriminator branch, we also evaluate the performance of adaptation for each class
by a class-conditional loss.

For source domain images, we employ the ground-truth class labels $Y_s$ to calculate the loss for each class and average over classes
to form a class-conditional binary cross entropy loss:
\begin{small}
\begin{equation}
\label{class_bceloss_source}
\mathcal{L}_{cbce\_ s} (U_s, Y_s, l_d)=
\frac{1}{C}\sum_{c=1}^C\left(
\frac {\sum_{h, w} Y_s^{(h, w, c)} \mathcal{L}_{bce} (U_s^{(h, w)}, l_d)} 
{\sum_{h, w} Y_s^{(h, w, c)} + \epsilon}
\right)
\end{equation} 
\end{small}
where the ground truth domain label $l_d$ is set to $l_d = 0$ when training the discriminator $\mathbf D$,
and to $l_d = 1$ when training the encoder $\mathbf E$ and segmentation module $\mathbf S$, to confuse the discriminator.
For target domain images, we do not have ground-truth class labels, and so we employ 
the pixel-level class predictions $P_t$ instead
to form a pseudo-label $\hat{Y}_t$ by selecting the class with the highest prediction value:
\begin{equation}
\label{pseudo_label_for_target1}
\hat{Y}_t^{h, w, c} = \left\{
             \begin{array}{lcl}
             &{1} &\text{if} \ \ {c = \mathop{\arg\max} P_t^{h,w}} \\
             &{0} &\text{otherwise}  
             \end{array}  
        \right. 
\end{equation} 

The above pixel-level class predictions will not be confident since the segmentation network is highly rely on source domain images.
It can be used as an indication that domain shift is interfering with classification.
Therefore, it is necessary to focus more on the adaptation of these uncertain regions by giving them a large weight.
We identify these ambiguous pixels by a label $N_t$: 
\begin{equation}
\label{pseudo_label_for_target2}
N_t^{h, w} = \left\{
             \begin{array}{lcl}
             &{1} &\text{if} \ \ {\mathop{\max}_c P_t^{h,w,c} < th_n} \\
             &{0} &\text{otherwise}  
             \end{array}  
        \right. 
\end{equation}  
where $th_n$ is a threshold constant for selecting the uncertain pixels. We then add an additional term to the fine-scale domain adaptation loss that will serve to upweight these regions during feature alignment. 
The final class-conditional binary cross entropy loss for target domain becomes:
{\setlength\abovedisplayskip{1pt}
\begin{small}
\begin{equation}
\begin{aligned}
\label{class_bceloss_target}
\mathcal{L}_{cbce_ t} (U_t, \hat Y_t, l_d)=
&
\frac{1}{C}\sum_{c=1}^C
\left( 
\frac {\sum_{h, w} \hat Y_t^{(h, w, c)} \mathcal{L}_{bce} (U_t^{(h, w)}, l_d)}
{\sum_{h, w} \hat Y_t^{(h, w, c)} + \epsilon}
\right)\\
+  
&
\lambda_n  \frac {\sum_{h, w} N_t^{(h, w)} \mathcal{L}_{bce} (U_t^{(h, w)}, l_d)} 
{\sum_{h, w} N_t^{(h, w)} + \epsilon}
\end{aligned}
\end{equation}
\end{small}
\setlength\belowdisplayskip{-1pt}}

The fine-scale class-conditional adaptation discriminator loss over all images is then:
\begin{equation}
\begin{aligned}
\label{Dloss2}
\mathcal{L}_{D2}(\mathbf{D}) = \  & \lambda_s \mathcal{L}_{cbce\_ s}(U_s, Y_s, 0) \\
+ & \lambda_t \mathcal{L}_{cbce\_ t}(U_t, \hat Y_t, 1)
\end{aligned}
\end{equation} 

The generative component of the adversarial fine-scale loss 
trained on feature encoder $\mathbf E$ and segmentation decoder $\mathbf S$
is defined symmetrically:
\begin{equation}
\begin{aligned}
\label{advloss2}
\mathcal{L}_{adv2}(\mathbf{E}, \mathbf{S}) = \  & \lambda_s \mathcal{L}_{cbce\_ s}(U_s, Y_s, 1) \\
+ & \lambda_t \mathcal{L}_{cbce\_ t}(U_t, \hat Y_t, 0)
\end{aligned}
\end{equation} 

For stability, we blend these class-conditional fine-scale losses with the conventional losses from the basic architecture
defined in Equations \ref{Dloss1} and \ref{advloss1}:
\begin{equation}
\label{fine_disloss}
\mathcal{L}_{D\_ fine}(\mathbf D) = \beta \mathcal{L}_{D1} + (1 - \beta) \mathcal{L}_{D2}
\end{equation}
\begin{equation}
\label{fine_advloss}
\mathcal{L}_{adv\_ fine}(\mathbf E, \mathbf S) = \beta \mathcal{L}_{adv1} + (1 - \beta) \mathcal{L}_{adv2}
\end{equation}

By utilizing the class-conditional loss, we manage to evaluate the performance among all classes equally for
our fine-scale discriminator branch to further improve the performance on less frequent classes.

\subsection{Class-Conditional Segmentation Loss} \label{section: Class-based Loss for Segmentation}

The conventional loss employed for pixel-level semantic segmentation is the pixel-level
cross-entropy loss.  
The final value of segmentation loss will be the mean loss of all pixels regardless of the class information.
Unfortunately, this has the drawback that classes that are less frequent at the
pixel-level, either because regions of that class occur infrequently or because they tend to be small, 
do not contribute substantially to the loss function,
while the pixels belongs to classes which are high frequent dominate the training process.
And thus performance of the trained system
for the less frequent classes can be poor,
since the pixels belongs to less frequent classes tend to be neglect during the training process. 
For domain adaptation systems, this has the additional consequence
that the system may never learn how to align representations across domains for these infrequent classes.

To begin to address this problem, we introduce a modified class-conditional loss for segmentation, which will
serve to distribute the loss more evenly across classes.  In particular, we employ a blend of the
dice loss~\cite{milletari2016v} and the cross-entropy loss to train our segmentation network.
The idea of dice loss comes from the dice coefficients,
and has been widely used in medical image segmentation~\cite{nie2018asdnet, wong20183d}. It has the form:
\begin{equation}
\label{diceloss1}
\mathcal{L}_{dice}(\mathbf E, \mathbf S) = 1 -
\frac{1}{C}\sum_{c=1}^C\left(
\frac {2 \sum_{h, w} Y_s^{(h, w, c)} P_s^{(h, w, c)}} 
{\sum_{h,w}(Y_s^{(h, w, c)} + P_s^{(h, w, c)}) + \epsilon}
\right)
\end{equation}

Note that the loss is formed as the complement of a normalized segmentation performance averaged
over classes.  The normalization, similar in spirit to intersection-over-union, will roughly equalize 
the contribution of each class to the loss function.  $\epsilon$ is a small constant that prevents division
by 0 for classes that do not appear in ground truth or predictions within an image.  

Since the up-weighting of rare classes may introduce instability in training, we elect to employ a blend of
the dice loss with the conventional cross-entropy loss  (Equation (\ref{segloss1})) to form the segmentation
prediction loss $\mathcal{L}_{pred}$:
\begin{equation}
\label{predloss}
\mathcal{L}_{pred}(\mathbf E, \mathbf S) = \alpha \mathcal{L}_{seg} + (1- \alpha) \mathcal{L}_{dice}
\end{equation}

By this class-conditional loss for segmentation, we manage to evaluate the performance of prediction among 
all classes equally on source domain images, which further improve the performance of segmentation prediction
for target domain images after the adaptation.

\subsection{Complete Training Loss}
To summarize, the complete training process combines the class-conditional segmentation loss (Equation
\ref{predloss}),  fine- and coarse-scale class-conditional domain adaptation discriminator losses 
(Equations \ref{fine_disloss} and \ref{Dloss_class}) and fine- and coarse-scale domain adaptation adversarial losses (Equation \ref{fine_advloss} and \ref{advloss_class}):
\begin{equation}
\label{total_D}
\mathop{\min}_\mathbf{D} \mathcal{L}_{D\_ fine}  +  \mathcal{L}_{D\_ coarse} 
\end{equation}
\begin{equation}
\label{total_G}
\mathop{\min}_{\mathbf{E}, \mathbf{S}} \  \mathcal{L}_{pred}
+ \mathcal{L}_{adv \_fine} + \mathcal{L}_{adv \_ coarse} 
\end{equation}

\begin{table*}[ht]
\centering
\resizebox{\textwidth}{!}{%
\begin{tabular}{c|c|cccccccccccccccccccc}
\toprule
\multicolumn{22}{c}{\textbf {GTA5 $\rightarrow$ Cityscapes}}                                                                                                                                           \\ \hline
 &  & 1 & 5 & 2 & 11 & 10 & 7 & 17 & 12 & 3 & 9 & 6 & 8 &  18 & 4 & 15 & 14 & 16 & 19 & 13 & \\
 & \rotatebox{90} { Meth.}  & \rotatebox{90}{ road} & \rotatebox{90}{ side.} & \rotatebox{90}{ buil.} & \rotatebox{90}{ wall} & \rotatebox{90}{ fence} & \rotatebox{90}{ pole} & \rotatebox{90}{ light} & \rotatebox{90}{ sign} & \rotatebox{90}{ vege.}  & \rotatebox{90}{ terr.} & \rotatebox{90}{ sky}  & \rotatebox{90}{ pers.} & \rotatebox{90}{ rider} & \rotatebox{90}{ car}  & \rotatebox{90}{ truck} & \rotatebox{90}{ bus}  & \rotatebox{90}{ train} & \rotatebox{90}{ mbike.} & \rotatebox{90}{ bike} & \rotatebox{90} { mIoU} \\ 
\hline 
 CDA~\cite{zhang2019curriculum}  & \multirow{1}{*}{CT}
& 72.9 & 30.0 & 74.9 & 12.1 & 13.2 & 15.3 & 16.8 & 14.1 & 79.3 & 14.5 & \textbf{75.5} & 35.7 & 10.0 & 62.1 & 20.6 & 19.0 & 0.0 & \textbf{19.3} & 12.0 & 31.4 \\
CBST-SP~\cite{zou2018unsupervised} &  \multirow{1}{*}{ST}
& \textbf{90.4} & 50.8 & 72.0 & 18.3 & 9.5 & \textbf{27.2} & \textbf{28.6} & 14.1 & \textbf{82.4} & 25.1 & 70.8 & 42.6 & \textbf{14.5} & 76.9 & 5.9 & 12.5 & \textbf{1.2} & 14.0 & \textbf{28.6} & 36.1 \\ 
Ours &  & 90.0&  \textbf{36.2}&  \textbf{79.1}&  \textbf{25.0}&  \textbf{18.9}&  26.8&  27.6&  \textbf{16.5}&  80.8&  \textbf{31.1}&  73.4&  \textbf{48.4}&  12.8&  \textbf{81.2}&  \textbf{25.6}&  \textbf{24.8}&  0.0&  12.5&  5.4& \textbf{37.7}\\
\hline
AdaptSeg~\cite{tsai2018learning}  & \multirow{3}{*} {AT-P}
& 87.3 & 29.8 & 78.6 & 21.1 & 18.2 & 22.5 & 21.5 & 11.0 & 79.7 & 29.6 & 71.3 & 46.8 & 6.5 & 80.1 & 23.0 & 26.9 & 0.0 & 10.6 & 0.3 & 35.0 \\
ADVENT~\cite{vu2019advent}  &
& 86.9 & 28.7 & 78.7 & \textbf{28.5} & \textbf{25.2} & 17.1 & 20.3 & 10.9 & 80.0 & 26.4 & 70.2 & 47.1 & 8.4 & \textbf{81.5} & 26.0 & 17.2 & \textbf{18.9} & 11.7 & 1.6 & 36.1 \\
CLAN~\cite{luo2019taking}  & 
& 88.0 & 30.6 & \textbf{79.2} & 23.4 & 20.5 & 26.1 & 23.0 & 14.8 & \textbf{81.6} & \textbf{34.5} & 72.0 & 45.8 & 7.9 & 80.5 & \textbf{26.6} & \textbf{29.9} & 0.0 & 10.7 & 0.0 & 36.6\\
\hdashline 
FCNs in the Wild~\cite{hoffman2016fcns}  & \multirow{3}{*} {AT-F}
& 70.4 & 32.4 & 62.1 & 14.9 & 5.4 & 10.9 & 14.2 & 2.7 & 79.2 & 21.3 & 64.6 & 44.1 & 4.2 & 70.4 & 8.0 & 7.3 & 0.0 & 3.5 & 0.0 & 27.1 \\
SIBIN~\cite{luo2019significance} &
& 83.4 & 13.0 & 77.8 & 20.4 & 17.5 & 24.6 & 22.8 & 9.6 & 81.3 & 29.6 & \textbf{77.3} & 42.7 & 10.9 & 76.0 & 22.8 & 17.9 & 5.7 & \textbf{14.2} & 2.0 & 34.2 \\
Ours &  &\textbf{90.0}&  \textbf{36.2}&  79.1&  25.0&  18.9&  \textbf{26.8}&  \textbf{27.6}&  \textbf{16.5}&  80.8&  31.1&  73.4&  \textbf{48.4}&  \textbf{12.8}&  81.2&  25.6&  24.8&  0.0&  12.5&  \textbf{5.4}& \textbf{37.7}\\
\bottomrule
\end{tabular}%
}
\setlength{\belowcaptionskip}{10pt}\centering\caption{Adaptation from GTA5 to Cityscapes.
We present the per-class IoU and mean IoU. 
The numbers above all classes are the indexes of their frequency in a descending order based on Cityscapes. 
(Please refer to~\cite{cordts2016cityscapes} for more details)
"CT", "ST" and "AT" represent curriculum-learning method, self-training.
and adversarial-learning method.
"P" and "F" represent prediction-level adaptation and feature-level adaptation.
We highlight the best result in each column in \textbf{bold}.}
\label{Tab: gta52cityscape}
\end{table*}

\section{Experiments}

In this section, we evaluate our class-conditional domain adaptation method and present the experimental results.
First, we introduce the used datasets and some implementation details of our network architecture.
Then, we show the comparison with the state-of-art method and discuss the effectiveness for our CCDA method with the ablation study.

\begin{table*}[ht]
\centering
\resizebox{\textwidth}{!}{%
\begin{tabular}{c|c|cccccccccccccccccc}
\toprule
\multicolumn{20}{c}{\textbf {SYNTHIA $\rightarrow$ Cityscapes}}  \\ 
\hline 
& & 1 & 5 & 2 & 10 & 9 & 7 & 14 & 11 & 3 & 6 & 8 & 15 & 4 & 13 & 16 & 12 & \\
& \rotatebox{90} { Meth. }       & \rotatebox{90} { road } & \rotatebox{90} { side. } & \rotatebox{90} { buil. } & \rotatebox{90} { wall* } & \rotatebox{90} { fence* } & \rotatebox{90} { pole* } & \rotatebox{90} { light } & \rotatebox{90} { sign } & \rotatebox{90} { vege. } & \rotatebox{90} { sky } & \rotatebox{90} { pers. } & \rotatebox{90} { rider } & \rotatebox{90} { car } & \rotatebox{90} { bus } & \rotatebox{90} { mbike. } & \rotatebox{90} { bike } & \rotatebox{90} { mIoU }  & \rotatebox{90} { mIoU* }  \\ 
\hline
CDA~\cite{zhang2019curriculum}  &\multirow{1}{*}{CT}
& 57.4 & 23.1 & 74.7 & 0.5 & \textbf{0.6} & 14.0 & 5.3 & 4.3 & 77.8 & 73.7 & 45.0 & 11.0 & 44.8 & \textbf{21.2} & 1.9 & 20.3 & 29.7 & 35.4 \\
CBST-SP~\cite{zou2018unsupervised}  & \multirow{1}{*}{ST}
& 69.6 & 28.7 & 69.5 & \textbf{12.1} & 0.1 & \textbf{25.4} & \textbf{11.9} & \textbf{13.6} & \textbf{82.0} & \textbf{81.9} & \textbf{49.1} & \textbf{14.5} & 66.0 & 6.6 & 3.7 & \textbf{32.4} & \textbf{35.4} & 36.1 \\
Ours &  & \textbf{82.6} &  \textbf{34.2} &  \textbf{76.9} &  2.6 &  0.2 &  23.8 &  3.5 &  7.7 & 77.9  & 79.5 & 44.2 & 8.2 &  \textbf{73.4} &  20.9 & \textbf{4.0} & 14.2 & 34.6 & \textbf{40.6} \\
\hline
AdaptSeg~\cite{tsai2018learning}  &\multirow{3}{*} {AT-P}
& 78.9 & 29.2 & 75.5 & - & - & - & 0.1 & 4.8 & 72.6 & 76.7 & 43.4 & 8.8 & 71.1 & 16.0 & 3.6 & 8.4 & - & 37.6 \\
ADVENT~\cite{vu2019advent}  &
& 67.9 & 29.4 & 71.9 & \textbf{6.3} & \textbf{0.3} & 19.9 & 0.6 & 2.6 & 74.9 & 74.9 & 35.4 & \textbf{9.6} & 67.8 & \textbf{21.4} & \textbf{4.1} & \textbf{15.5} & 31.4 & 36.6 \\
CLAN~\cite{luo2019taking} &
& 80.4 & 30.7 & 74.7 & - & - & - & 1.4 & 8.0 & 77.1 & 79.0 & 46.5 & 8.9 & \textbf{73.8} & 18.2 & 2.2 & 9.9 & - & 39.3 \\
\hdashline 
FCNs in the Wild~\cite{hoffman2016fcns} &\multirow{4}{*} {AT-F}
& 11.5 & 19.6 & 30.8 & 4.4 & 0.0 & 20.3 & 0.1 & \textbf{11.7} & 42.3 & 68.7 & \textbf{51.2} & 3.8 & 54.0 & 3.2 & 0.2 & 0.6 & 20.2 & 22.9 \\
Cross-city~\cite{chen2017no} &
& 62.7 & 25.6 & 78.3 & - & - & - & 1.2 & 5.4 & \textbf{81.3} & \textbf{81.0} & 37.4 & 6.4 & 63.5 & 16.1 & 1.2 & 4.6 & - & 35.7 \\
SIBIN~\cite{luo2019significance}  &
& 70.1 & 25.7 & \textbf{80.9} & - & - & - & \textbf{3.8} & 7.2 & 72.3 & 80.5 & 43.3 & 5.0 & 73.3 & 16.0 & 1.7 & 3.6 & - & 37.2 \\
Ours &  & \textbf{82.6} &  \textbf{34.2} &  76.9 &  2.6 &  0.2 &  \textbf{23.8} &  3.5 &  7.7 & 77.9  & 79.5 & 44.2 & 8.2 &  73.4 &  20.9 & 4.0 & 14.2 & \textbf{34.6} & \textbf{40.6} \\
\bottomrule
\end{tabular}%
}
\setlength{\belowcaptionskip}{10pt}\centering\caption{Adaptation from SYNTHIA to Cityscapes.
The table setting is the same as Table \ref{Tab: gta52cityscape},
while mIoU and mIoU* are averaged over 16 and 13 categories, respectively.}
\label{Tab: synt2cityscape}
\end{table*}

\begin{table}[hbp]
\centering
\resizebox{\textwidth}{!}{%
\begin{tabular}{c|ccc|c}
\toprule
\multirow{2}{*} {Task} & \multicolumn{3}{c}{Method} &  \multirow{2}{*} {mIoU} 
\\
& $\mathcal{L}_{basic}$ & $\mathcal{L}_{basic\_class}$ & $\mathcal{L}_{coarse}$  
& \\
\hline 
\multirow{3}{*} 
{\begin{tabular}[c]{@{}c@{}}GTA5 $\rightarrow$ \\ Cityscapes \end{tabular}}
& \checkmark &  &  & 34.9\\
& \checkmark &   \checkmark &   & 37.0\\
& \checkmark & \checkmark  &  \checkmark &  37.7\\
\bottomrule 
\end{tabular}%
}
\setlength{\belowcaptionskip}{10pt}\centering\caption{Ablation Study on our CCDA mehtod.}
\label{Tab: ablation_r}
\end{table}

\begin{figure*}[ht]
\centering
\includegraphics[width=0.8\textwidth, height=0.42\textheight]{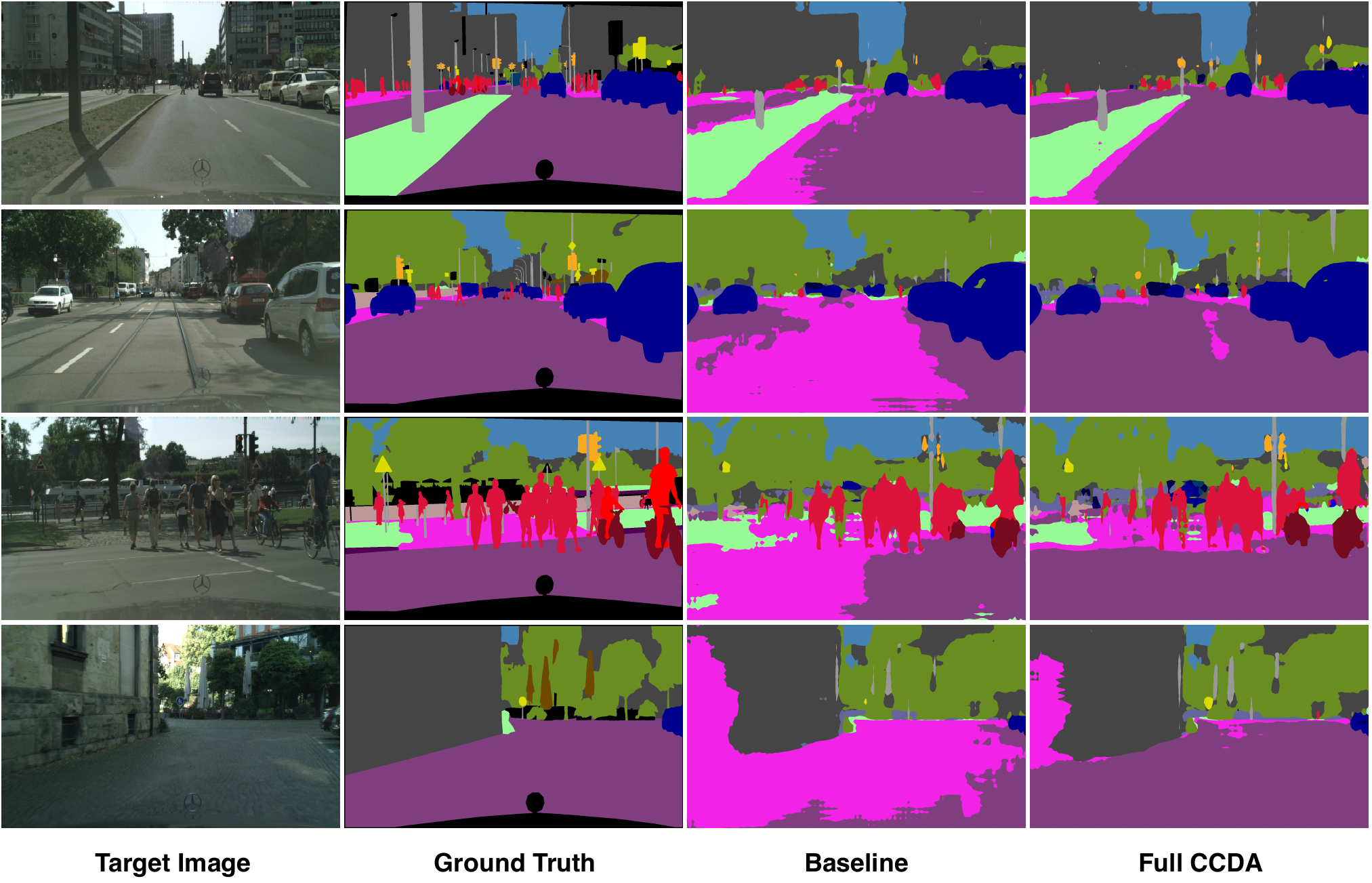}
 \caption{Example results of our proposed Class-Conditional Domain Adaptation (CCDA).
 for GTA5 $\rightarrow$  Cityscapes task. 
 For each target image, we show the corresponding Ground-truth map, 
 the result for the baseline CCDA architecture in Section \ref{section: Basic Domain Adaptation Structure},
 and the result for our full CDDA system.}
\label{fig: show}
\end{figure*}

\subsection{Datasets}
Manual generation of pixel-level ground truth for semantic segmentation is expensive.
Since for synthetic datasets, pixel-level labels can be derived directly from the generative
model, synthetic data has the potential to vastly expand the quantity of labeled data used
to train deep semantic segmentation networks.  But translating this into greater accuracy
at inference time requires bridging any domain shift between the synthetic training data and the
real test data.

With this motivation, we evaluate our class-conditional domain adaptation method by employing two synthetic source domain datasets (SYNTHIA~\cite{ros2016synthia} and GTA5~\cite{richter2016playing}) and a real-world target domain dataset (Cityscapes~\cite{cordts2016cityscapes}).
This defines two adaptation tasks: GTA5 $\rightarrow$ Cityscapes, and SYNTHIA $\rightarrow$ Cityscapes.

The GTA5 dataset is a large synthetic dataset with 24,966 images with pixel-level ground-truth labels.
The image resolution is 1914 $\times$ 1052 pixels and the class labels are compatible with the Cityscapes dataset.  
The SYNTHIA dataset is also a widely used synthetic dataset for domain adaptation, which contains 9,400 images with pixel-level ground-truth labels. The image resolution is 960 $\times$ 720 pixels.  
For the real-world images, Cityscapes dataset comprises 2,975 training images 
and 500 validation images with the resolution of 2048 $\times$ 1024 pixels.

To train our domain adaptation model,  
we employ both the images and the ground truth labels from either the GTA5 dataset or SYNTHIA datasets for the source domain, 
and only the images (not the labels) from the Cityscapes training set for the target domain.
We evaluate our model on the Cityscapes validation set, over 19 classes for GTA5 $\rightarrow$ Cityscapes
adaptation and over 13 and 16 classes for SYNTHIA $\rightarrow$ Cityscapes, as per convention 
like in~\cite{tsai2018learning, zhang2019curriculum}. 


\subsection{Implementation Details}

We apply PyTorch for our implementation using a single GeForce RTX 2080 Ti GPU with 11 GB memory.
For segmentation, we use the DeepLab-v2~\cite{chen2018deeplab} framework with a small
VGG16~\cite{simonyan2014very} pre-trained model as backbone for our feature encoder $\mathbf E$ and
segmentation decoder $\mathbf S$.
For discriminator module $\mathbf D$, the fine-scale branch has a structure similar to~\cite{tsai2018learning, luo2019taking},
consistsing of 5 convolution layers with channel numbers $\{64, 128, 256, 512, 1 \}$.
To preserve fine-scale detail, we use 3 $\times$ 3 kernels and a stride of 1.
A final up-sampling layer is added at the end of this branch to rescale the output to the input image resolution.
For the coarse-scale branch, we share the first two convolution layers with the fine-scale branch,
and then apply 3 convolution layers with channel numbers $ \{256, 512, C\times 2 \}$ with kernel 3 and stride of 2
for downsampling.
Except for the last convolution layer in both branches,
each convolution layer in our discriminator module is followed by a Leaky-ReLU~\cite{maas2013rectifier} with a slope of $0.2$ for negative inputs.

To train our feature encoder $\mathbf E$ and segmentation decoder $\mathbf S$,
we use the Stochastic Gradient Descent (SGD) optimizer~\cite{bottou2010large} with the momentum of 0.9
and the weight decay is 5$e -$4.
The initial learning rate is set to 2.5$e -$4 and decays during training.
For discriminator module $\mathbf D$, we apply ADAM~\cite{kingma2014adam} optimizer
with $\beta 1 = 0.9$ and $\beta 2 = 0.99$.
The initial learning rate is set to 1$e -$4 and decayed with the same policy as SGD.
We train our model with a crop of 512 $\times$ 1024 with one source domain image and one target domain image at a time
the same as in~\cite{luo2019taking, luo2019significance}.
We set hyper-parameters $\lambda_s = \lambda_t = 0.0003$ for both fine-scale and coarse-scale branch.

\subsection{Results}
Table \ref{Tab: gta52cityscape} and Table \ref{Tab: synt2cityscape} summarizes the performance of our method compared with the state-of-the-art on the two transfer tasks GTA5 $\rightarrow$ Cityscapes,
and SYNTHIA $\rightarrow$ Cityscapes, respectively.
For a fair comparison, we choose several state-of-art methods using the same VGG16 as backbone as our method,
which include adversarial-learning methods with prediction-level adaptation AdaptSeg~\cite{tsai2018learning}, ADVENT~\cite{vu2019advent},
CLAN~\cite{luo2019taking};
and adversarial-learning methods with feature-level adaptation FCNs in the Wild~\cite{hoffman2016fcns}, Cross-city~\cite{chen2017no},
SIBIN~\cite{luo2019significance}.
We do not compare with the two latest state-of-art method~\cite{du2019ssf, tsai2019domain},
since they use different setting of Cityscape dataset or add extra pixel-level adaptation (image style-transfer) module. 
We also compare our method with self-training method CBST-SP~\cite{zou2018unsupervised};
curriculum-learning method CDA~\cite{zhang2019curriculum}.
These two methods achieve better performance on lower frequency classes due to their strategy of alternate selection
of target domain samples for training the segmentation.

In Table \ref{Tab: gta52cityscape}, 
we present our experimental results on the GTA5 $\rightarrow$ Cityscapes task
compared with the chosen state-of-art methods.
This table shows our proposed CCDA method
performs better on average than all of these methods, and this advantage derives from improvements over a wide range of classes.
We observe that our class-conditional method boosts the performance of lower-frequency classes substantially
while maintaining performance for higher-frequency classes like road, building, vegetation, car, 
and thus ultimately a higher mean IoU performance. 

In Table \ref{Tab: synt2cityscape}, comparison with current state-of-the-art methods on SYNTHIA $\rightarrow$ Cityscapes transfer task 
shows that our CCDA method performs favorably against the other algorithms on mIoU, which indicates that our method increases the overall performance.
Especially, compared with other adversarial learning methods, our CCDA has advantages on lower-frequency classes.
While for self-training and curriculum-learning method which perform better on several least frequent classes,
we can still reach a comparable results on these classes and outperform them on the overall performance.

\subsection{Ablation Studies}
To better understand the impact of each component of our adaptation model,
we conducted an ablation study by selectively deactivating each component
and measuring the effect on performance for the GTA5 $\rightarrow$ Cityscapes transfer task.
In particular, we define three nested subset models:

1) $\mathcal{L}_{basic}$: using the basic domain adaptation architecture in  
Section \ref{section: Basic Domain Adaptation Structure} with the segmentation loss and a fine-scale basic discriminator.

2) $\mathcal{L}_{basic} + \mathcal{L}_{basic\_class}$:
adding class-conditional loss for segmentation in Section \ref{section: Class-based Loss for Segmentation} as well as the class-conditional loss for fine-scale discriminator in Section \ref{section: Class-based Loss for Adaptation} on the basic architecture.

3) $\mathcal{L}_{basic} + \mathcal{L}_{class\_ based} + \mathcal{L}_{coarse}$:
further adding the class-conditional coarse-scale branch for discriminator in 
Section \ref{section: Class-conditional Coarse-scale Branch}.

The result is showed in Table \ref{Tab: ablation_r}.  Our class-conditional loss on the basic architecture
(both segmentation and fine-scale discriminator)
gains 2.1\% improvements together
and the designed coarse-scale branch brings another 0.7\% improvements. 
It verifies the effectiveness of our CCDA method, 
including both class-based loss and the class-based coarse-scale branch.   
We also present some qualitative segmentation examples in Figure \ref{fig: show}. 
This figure also verifies the effectiveness of our CCDA method.
The performance of our CCDA method outperforms the baseline
structure in two ways.
Firstly, it provides a cleaner and more accurate prediction on higher-frequency like
road and sidewalk.
Secondly, it improves the performance of lower-frequency classes like light and sign.

\section{Conclusions}
We have developed a novel approach to solving an important problem in domain adaptation for semantic segmentation, namely, the poor performance often observed for infrequent classes.  The solution hinges on the  introduction of class-conditioning at multiple points in the model, including segmentation, coarse-scale domain adaptation and fine-scale domain adaptation, and upon equalizing across classes at several stages in the computation.  
Evaluation on two transfer tasks demonstrates the effectiveness of our method, which boosts performance on infrequent classes while maintaining performance on the remaining classes. Generally, the proposed class-conditional domain adaptation method outperforms the state of the art on average, due to superior performance on a broad range of classes. 
\\ \hspace*{\fill} \\
\noindent \textbf{Acknowledgements: }
We would like to thank the York University
Vision:  Science to Applications (VISTA) program for its support.

{\small
\bibliographystyle{ieee_fullname}
\bibliography{egbib}
}

\end{document}